\documentclass[10pt,journal,compsoc]{IEEEtran}
\usepackage{amsfonts}
\usepackage{bm}
\usepackage{multirow}
\usepackage{amssymb}
\usepackage{amsmath}
\usepackage{makecell}

\ifCLASSOPTIONcompsoc
  \usepackage[nocompress]{cite}
\else
  \usepackage{cite}
\fi
\ifCLASSINFOpdf
   \usepackage[pdftex]{graphicx}
   \graphicspath{{../pdf/}{../jpeg/}}
   \DeclareGraphicsExtensions{.pdf,.jpeg,.png}
\else
   \usepackage[dvips]{graphicx}
   \graphicspath{{../eps/}}
   \DeclareGraphicsExtensions{.eps}
\fi
\usepackage{algorithm}  
\usepackage{algorithmic}
\usepackage[T1]{fontenc}

\usepackage{array}

\ifCLASSOPTIONcompsoc
 \usepackage[caption=false,font=footnotesize,labelfont=sf,textfont=sf]{subfig}
\else
 \usepackage[caption=false,font=footnotesize]{subfig}
\fi

\ifCLASSOPTIONcaptionsoff
 \usepackage[nomarkers]{endfloat}
\let\MYoriglatexcaption\caption
\renewcommand{\caption}[2][\relax]{\MYoriglatexcaption[#2]{#2}}
\fi
\let\MYorigsubfloat\subfloat
\renewcommand{\subfloat}[2][\relax]{\MYorigsubfloat[]{#2}}
\usepackage{url}

\pdfoutput=1
\usepackage{cite}
\usepackage{hyperref}

\begin{document}
%
\title{Reinforced Contact Tracing and Epidemic Intervention}
%
%
%
%

\author{~\IEEEmembership{Tao~Feng,
    Sirui~Song,
    Tong~Xia,
    Yong~Li}
    
    \IEEEcompsocitemizethanks{\IEEEcompsocthanksitem T. Feng, S. Song, T. Xia and Y. Li are with Beijing National Research Center for Information Science and Technology (BNRist), Department of Electronic Engineering, Tsinghua University, Beijing 100084, China. Email: liyong07@tsinghua.edu.cn.}
}

%
%

\markboth{IEEE TRANSACTIONS ON KNOWLEDGE AND DATA ENGINEERING}%
{Shell \MakeLowercase{\textit{et al.}}: Bare Demo of IEEEtran.cls for Computer Society Journals}
%



\IEEEtitleabstractindextext{%
\begin{abstract}
  The recent outbreak of COVID-19 poses a serious threat to people's lives. Epidemic control strategies have also caused damage to the economy by cutting off humans' daily commute. In this paper, we develop an Individual-based Reinforcement Learning Epidemic Control Agent (IDRLECA) to search for smart epidemic control strategies that can simultaneously minimize infections and the  cost of mobility intervention. IDRLECA first hires an infection probability model to calculate the current infection probability of each individual. Then, the infection probabilities together with individuals'  health status and movement information are fed to a novel GNN to estimate the spread of the virus through human contacts. The estimated risks are used to further support an RL agent to select individual-level epidemic-control actions. The training of IDRLECA is guided by a specially designed reward function considering both the cost of mobility intervention and the effectiveness of epidemic control. Moreover, we design a constraint for control-action selection that eases its difficulty and further improve exploring efficiency. Extensive experimental results demonstrate that IDRLECA can suppress infections at a very low level and retain more than $95\%$ of human mobility. 
\end{abstract}

}

\maketitle

\IEEEdisplaynontitleabstractindextext

%
\IEEEpeerreviewmaketitle

\section{Introduction}
The recent outbreak of COVID-19 has caused thousands of infections and deaths. Similar to most epidemics that can spread via human contact~\cite{balcan2010modeling}, control the spread of the COVID-19 virus requires cutting off human contacts. Governments have taken different epidemic-control strategies, such as travel-restriction orders, individual quarantine policies, and city lockdown~\cite{hale2020variation}. However, restricting human's daily mobility and gathering will inevitably pose a negative effect on economic growth. The current epidemic control strategies for COVID-19 has ultimately caused damage to the economy~\cite{Bonaccorsi15530,barua2020understanding}. 

To control the epidemic both efficiently and effectively, researchers have proposed smart and computational Epidemic-Prevention-and-Control (EPC) strategies in both group level and individual level. 
Group-level EPC strategies \cite{yang2020modified, song2020reinforced} aim to select customized epidemic-control actions for each population group. These works are mainly based on the SIR model \cite{SIR} which can characterize the development trend of the epidemic from a group-level view. However, Group-level EPC strategies ignore the unique situation of each individual, which may easily cause unnecessary 
mobility intervention costs or secondary transmission of infection. 
By contrast, individual-based EPC strategies exploit individual information to estimate infection risk for each individual, and further select a customized epidemic-control action for each individual \cite{rocha2016individual}. However, current individual-based EPC strategies\cite{c.elmohamed,kimexpert,donghierarchical,eubank2004modelling,watts1998collective} lack a module to estimate the spread of the virus through complex contacts between individuals.
To achieve an efficient and effective EPC result, we in this paper aim to maximally make use of available information and design an individual-based EPC strategy that can both minimize the number of infections and the social cost of epidemic control. 

The main challenges of our research are three-fold. \emph{First}, primitive individual information can hardly reflect an individual's infection risk. For example, an asymptomatic patient who has a very high infection risk is usually hard to detect just through symptoms. In other words, the large population and their complex information form a vast state space for control, making it very hard to extract effective information to support the selection for control actions.
\emph{Second}, the large and complex action space exacerbates the difficulty of control-action selection. If there exist $M$ people and $d$ kinds of control actions, the action space is $M^{d}$, which is growing exponentially. 
\emph{Third}, searching for a policy that achieves the dual objective of minimizing both infections and the social cost of implementing the strategy is hard. The two optimization goals will influence each other. For example, better control of the epidemic requires greater control efforts,  which will naturally increase mobility intervention costs. 

To solve the above challenges, we propose an Individual-based Reinforcement Learning Epidemic Control Agent (IDRLECA) by combining Graph Neural Network and Reinforcement Learning approach. Specifically, to deal with the vast-state-space challenge, we design an infection probability model to calculate the current infection probability of each individual, whose result is further added to the individual's state as auxiliary information. In order to better extract individual features hidden in his/her daily commute, we design a novel GNN which inputs with individuals’ states their visiting history and estimates their infection risks of individuals. 
As for the large-action-space challenge, we design and impose a constraint to control-action selection by requiring individuals with larger calculated infection probability should receive more stringent control actions.  
In response to the dual-objective optimization challenge, we carefully design a reward function considering both the social cost of EPC and the effectiveness of infection suppression. More importantly, the reward function is able to efficiently guide training. 

We build a simulation environment based on the PAPW Challenge\footnote{PAPW 2020: https://prescriptive-analytics.github.io/.}. and experimentally compare the performance of expert EPC strategies, winners in the PAPW Challenge, and our proposed IDRLECA. Extensive results show that IDRLECA achieves the best performance for both infection-suppression and mobility-retaining in all three compared scenarios.

In summary, this paper makes the following contributions:

\begin{itemize}
    \item We propose IDRLECA to minimize the number of infections and the social cost of EPC. IDRLECA achieves the best performance in both infection-suppression and mobility-retaining compared with expert baselines and PAPW winners.

    \item We propose a method to address the vast-state-space problem in individual-based EPC. Our method includes an infection-probability model and a novel GNN. 
    \item We design and impose a constraint to control-action selection to improve the exploration efficiency of IDRLECA in the extremely large action space.
\end{itemize}

The remainder of this paper is as follows. We first introduce our problem formulation in Section 2 and introduce our method in Section 3. The experiment results are presented in Section 4. We introduce the related works in Section 5 and conclude our paper in Section 6.
\section{Problem Formulation And Challenges}
In this section, we formulate the problem of individual-based EPC and discuss the challenges in finding an effective EPC policy. 

\subsection{Formulation}

We consider a within-city epidemic control scenario. The city is assumed to be composed of $N$ areas and has a population of $M$.
Each individual's health status can be: Susceptible, Asymptomatic, Symptomatic, and Recovered. Asymptomatic and Symptomatic individuals are both infected.
Each individual will commute between different areas according to some predefined commute rules. When people are staying in the same area, they have a probability to contact each other and their health status will change from Susceptible to Asymptomatic. The Asymptomatic status will Symptomatic after a predefined incubation time. Symptomatic individuals will be sent to the hospital and transit to Recover after a predefined time of treatment. 
Note that policymakers cannot distinguish between Susceptible individuals and Asymptomatic individuals. 
The goal of individual-based EPC is to select a control action for each individual in the Susceptible group and Asymptomatic group to minimize the number of infected people and the cost of intervention measures. Specifically, we define four kinds of control actions: No Intervention, Confine (no contact with people living outside his/her residential area), Quarantine (no stranger contact), Isolate (no contact). 
The above modeling for epidemic transmission and individual-based EPC actions is shown in Figure 1. 

\begin{figure}[ht]
    \centering
    \includegraphics[width=0.4\textwidth]{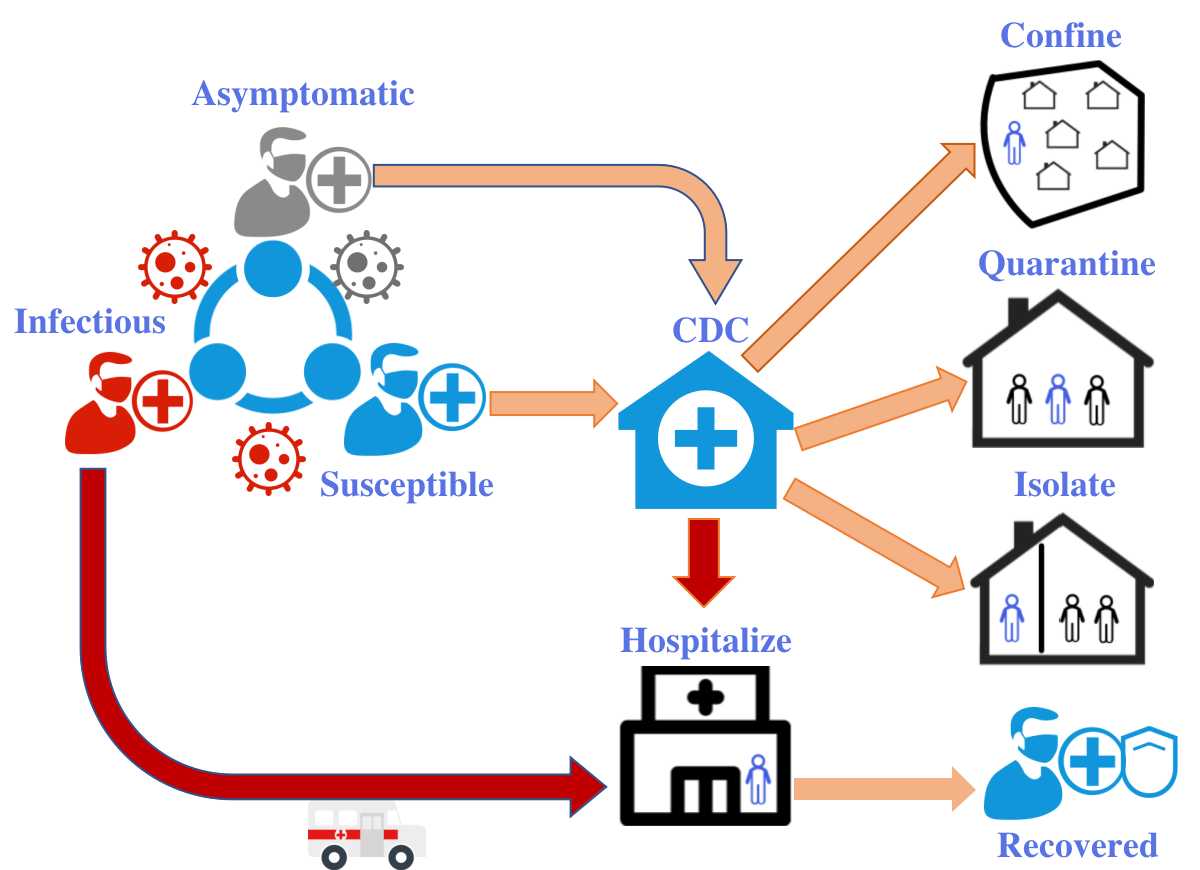}
    \caption{Epidemic Spread and Intervention(CDC: Center for Disease Control and prevention).}\label{fig:overview}
\end{figure}

Once the number of infected people exceeds a threshold, the medical system will be penetrated, leading to a rapid increase in medical costs. On the other hand, when the mobility intervention is greater than a certain threshold, the economic system will be paralyzed, also leading to a sharp increase in social cost. So we design the metric $Score$ to evaluate the total cost of an EPC policy to consider reducing the infections and maintaining the mobility  at the same time.
The smaller $Score$ value indicates better EPC results. $Score$ is defined as follows:
\begin{align}
&Q = \lambda_{h}*N_{h}+\lambda_{i}*N_{i}+\lambda_{q}*N_{q}+\lambda_{c}*N_{c}, \nonumber \\  
&Score = exp\left \{\frac{I}{\theta _{I}} \right \}+exp\left \{ \frac{Q}{\theta _{Q}} \right \}, \nonumber
\end{align}
where $I$ denotes  the total number of infected people within all simulation days, $Q$ denotes  the aggregate of mobility interventions, $N_{h}$, $N_{i}$, $N_{q}$ and $N_{c}$ denote  the accumulated number of hospitalized, isolated, quarantined, and confined people for all simulation days, $\theta _{I}$ and $\theta _{Q}$ refers to the soft thresholds for medical system's capacity and economic system's endurance.
$\lambda_{h}$, $\lambda_{i}$, $\lambda_{q}$ and $\lambda_{c}$ denote  scale factors.

In this paper, we aim to find an EPC policy that gives daily control actions for all individuals to minimize $Score$.

\subsection{Challenges}
Finding an effective EPC policy is challenging in three aspects:

    \subsubsection{Vast State Space} 
    The invisibility of asymptomatic patients and people's complex contacts makes the state space vast. It's difficult to extract effective features for control-action selection. To tackle this challenge, we propose two solutions. We design an infection probability model to calculate the current infection probability of each individual. The probability is added to the state of each individual as auxiliary information. Moreover, IDRLECA employs a novel GNN acquires the whole individuals’ state and the area visited history as input, which can estimate the infection risks through the contact between individuals. The estimated infection risks which measure the individual's ability to potentially infect others are further used as action thresholds to support the selection for actions.

    \subsubsection{Large Action Space} Individual-level epidemic control aims to select a control action for each individual, which brings an extremely large action space for this control problem. This further leads to low exploration efficiency for reinforcement learning. In order to solve this challenge, we design an infection probability model to calculate the current infection probability for each person and use IDRLECA to output different action control thresholds for each individual.  The estimated risks are further used to support RL's selection for action actions. 
    
    \subsubsection{Dual Objective Optimization} Since our goal is to minimize the social cost $Score$ which contains two optimization objectives of the entire epidemic control process. To solve this problem, we propose a special design instant reward, which considers the number of new infections on two consecutive days and the mobility intervention cost on the current day. 
    
\section{Methodology}

\begin{figure}[ht]
    \centering
    \includegraphics[width=0.48\textwidth]{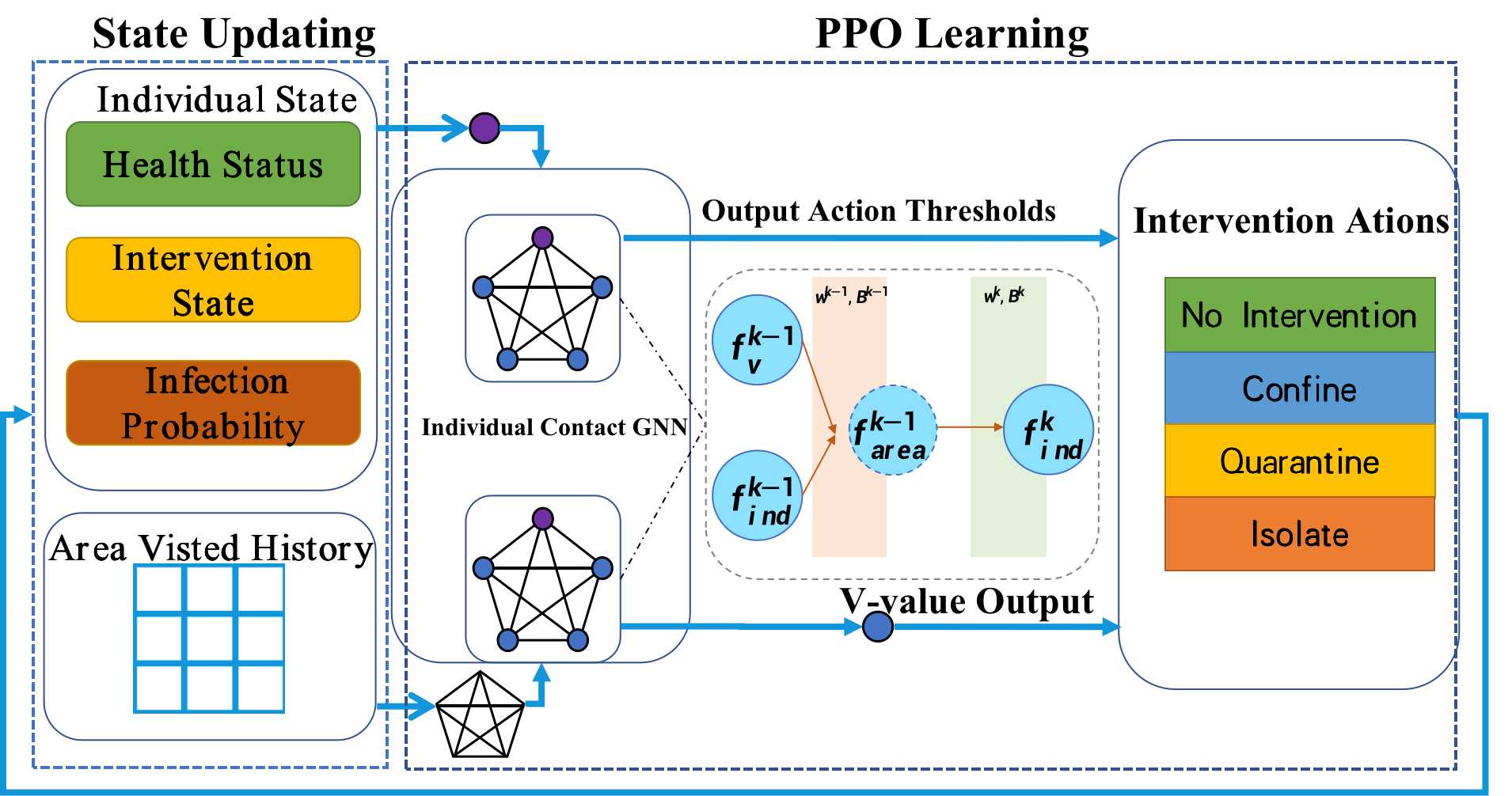}
    \caption{The detailed structure of proposed IDRLECA.}\label{fig:method}
\vspace{-0.2cm}
\end{figure}

To tackle the above challenges, we develop an Individual-based  Reinforcement Learning Epidemic Control Agent (IDRLECA) that employs a novel GNN and RL approach to search for smart control policies.  An overview of IDRLECA is shown in Figure \ref{fig:method}. 
At each time step, IDRLECA collects the health status, intervention state and area-visit-history for each individual and gives each a customized intervention action. In the rest of this section, we will provide the details of the IDRLECA.

\subsection{Infection Probability Model}
The difficulty of epidemic prevention and control lies in how to find asymptomatic infections and how to take timely and effective measures. To help the latter part of IDRLECA efficiently take use of effective information, we here design an infection probability model to  estimate the probability of an individual being infected. We define the probability of infection and health of the $i$-th person as $p_{i}^{infe}$ and $p_{i}^{hel}$, respectively. The infection probabilities of contacting with strangers and acquaintances are calculated by the simulation environment, denoted as $p_{s}$ and $p_{c}$, respectively. 
The infection probability model works as follows:

\textbf{Step 1:} Trace back all individuals' area-visit history in the past $T$ time steps.

\textbf{Step 2:} For individual $i, i = 1,2,...,M$, define his/her probability of being healthy as $p_{i,t}^{hel}$ at time step $t$. $p_{i,0}^{hel}$ is initialized to be 1 if individual $i$ is not infected. we have the following equation to update $p_{i,t}^{hel}$:
\begin{align}
    p_{i,t}^{hel} = p_{i, t-1}^{hel} \ast (1- p_s \frac{{N^{infe}_{t-1}}} {N^{area}_{t-1}}), t = 1,2,...,T,
\end{align}
where $N^{infe}_{t-1}$ and $N_{area}$ refer to the number of discovered infections and total number of visitors to the same area as individual $i$, respectively.

\textbf{Step 3:} Update $p_{i,T}^{hel}$ for acquaintances' contacts:
\begin{align}
    \hat{p}_{i,T}^{hel} = p_{i,T}^{hel} * (1-p_{c}).
\end{align}

\textbf{Step 4:} Acquire infection probability:
\begin{align}
    p_{i}^{infe} = 1- \hat{p}_{i, T}^{hel},
\end{align} 

After the above steps, we can obtain the estimated probability of an individual being infected. We will use it as auxiliary information and add it to each individual's state. Also, the estimated probabilities are used as prior knowledge for the agent selection control actions. Note that this estimation is not 100\% accurate since our estimation  simplifies the process of contact and spread of virus  between people. In the later part, we will combine GNN to solve this problem.

\subsection{Reinforcement Learning}
We propose IDRLECA to search smart strategies to minimize the spread of epidemic and cost of intervention at the same time. We treat all individuals in the area as one agent. Therefore, for IDRLECA, its status and actions are for all people. We use one day as the decision time interval. In the following, we will introduce our design of state, action and reward:

\begin{itemize}
    \item \textbf{State:} The state of IDRLECA is the integration of each individual's information, which is obtained at the start of one day. For each individual, the state includes infection state, intervention state, and the probability of infection calculated by Equation (1)$\sim$(3). 
    
    \item \textbf{Action:} The  action  at  each  step  for  the  agent  is  to  determine the intervention measure of each individual. The action contains no intervention, confine, and quarantine.  In order to ensure the flexibility of the policy,  we set the implementation time of actions to one day.

    \item \textbf{Reward:} The goal of our method is to minimize the total number of infected people, and to minimize the total intervention cost at the same time. Considering our dual objective optimization, we set the reward $r$  as follows,
    \begin{align}
    \label{equ:atten6}     
         r=-exp\left \{ \frac{\Delta I}{\theta _{I}} \right \}-exp\left \{ \frac{\Delta Q}{\theta _{Q}} \right \},
    \end{align}
    where $\Delta I$ and $\Delta Q$ denote the daily  incremental
    part of the  number  in the infected population between consecutive days and the cost of mobility intervention on the day, respectively.

    \item  \textbf{Learning Algorithm: } 
    IDRLECA employs a Proximal Policy Optimization(PPO)~\cite{schulman2017proximal} agent to find the optimal strategy that minimizes the number of infected people and the cost of prevention at the same time. The PPO agent adopts the actor-critic framework. The critic network is used to estimate the long-term reward of the action, and the actor network is to find the optimal action policy to achieve dual objective optimization. We also add an entropy bonus to ensure sufficient exploration when RL training~\cite{schulman2017proximal}.
\end{itemize}

\subsection{Individual Contact GNN} 
Since asymptomatic patients are indistinguishable, it's hard to trace all the contacts and infections caused by them. Moreover, vast modern traffic and complex social network structure make it more challenging to estimate the infection risk of each individual. To deal with this challenge, we propose a novel GNN, namely Individual Contact GNN, to estimate the infection risk of each individual.
Individual Contact GNN is used to build both the actor network and critic network in IDRLECA. The GNN regards individuals and city areas as two kinds of nodes. This enables us to model individual-individual contacts by individual-area-individual contacts, which further helps us to avoid the extremely large individual-individual contract matrix (size $M*M$). 

Specifically, Individual Contact GNN is designed on the basis of GraphSage ~\cite{hamilton2017inductive}. The state input to the GNN consists of health status, intervention state, infection probability for all individuals, and the edge-information inputs are the area-visit-history at different time steps.
We use $f_{area}^{k}, f_{ind}^{k}$ to denote for the area-nodes' features and individual-nodes' features outputted by the $k$-th GNN layer, respectively. 
The detailed layer-calculation of Individual Contact GNN is as follows:
\begin{align}
&f_{c}^{k-1}=softmax(f_{v}^{k-1}), \\
&f_{area}^{k-1}=\sigma(W^{k-1}(f_{c}^{k-1})^T f_{ind}^{k-1}+B^{k-1}), \\
&f_{ind}^{k}=\sigma (W^{k}f_{c}^{k-1}f_{area}^{k-1}+B^{k}),
\end{align}
where $f_{v}^{k-1}$ denotes for the area's visit history at the $k-1$ time step, $W^{k-1}, B^{k-1}, W^{k}, B^{k}$ denotes for trainable parameters.

 In the above equations, Equation (5)  uses the area-visit-history as edge weights; Equation (6)  aggregates weighted visitors' characteristics to calculate the area-node feature; Equation (7) aggregates the features of areas where an individual has visited to calculate individual-node feature.

\subsection{Constraint for Control-Action Selection} 
As discussed before, the EPC problem has a extremely large action space, which challenges policy search. To address this issue, we incorporate prior knowledge into the control-selection step. Specifically, we let the actor network of IDRLECA first outputs four values $<p_{i,1},p_{i,2},p_{i,3},p_{i,4}>$ for individual $i, i=1,2,3,...,M$. Then, we transform the four values to three thresholds:
\begin{align}
    P_{i,1} = \frac{e^{-p_{i,1}}}{e^{-p_{i,1}} + e^{-p_{i,2}} + e^{-p_{i,3}} + e^{-p_{i,4}}}, \\
    P_{i,3} = \frac{e^{-p_{i,1}} + e^{-p_{i,2}}}{e^{-p_{i,1}} + e^{-p_{i,2}} + e^{-p_{i,3}} + e^{-p_{i,4}}}, \\
    P_{i,2} = \frac{e^{-p_{i,1}}+ e^{-p_{i,2}} + e^{-p_{i,3}}}{e^{-p_{i,1}} + e^{-p_{i,2}} + e^{-p_{i,3}} + e^{-p_{i,4}}}.
\end{align}
Through the above equations, we can ensure $0\leq P_{i,1}\leq P_{i,2}\leq P_{i,3}\leq1$. Thus, $P_{i1}, P_{i2},P_{i3}$ can be used as different  infection risk levels, which  considers the risk of individual infection and  individual’s ability to potentially infect others. It's natural and reasonable to expect that an individual with a higher infection risk should receive a more stringent control action. The infection risk levels are further used as the  thresholds for the infection probabilities estimated in Section 2, which imposes a constraint that individuals with higher infection probability will have higher infection risk and  receive more stringent control actions. In this way, individuals with high probability of infection are not identified as low risk, thus reducing unnecessary strategy exploration.

By comparing the pre-calculated infection probability $p_i^{infc}$ with $<P_{i1}, P_{i2},P_{i3}>$, we define the action-selection rule in Table 1. It can be seen from Table 1 that as the infection probability goes from low to high, the corresponding intervention actions become more and more stringent. There are different thresholds for different individuals, which fully takes into account the differences in individual states. 

\begin{table}[h]
    \centering
    \caption{ Action-Selection Rule. }
    \begin{tabular}{c|c}
    \Xhline{1pt}
    Infection probability& Intervention actions \\
    \Xhline{1pt}
    $0\leq p_{i}^{infc} \leq P_{i1}$&No intervention\\
    $P_{i1} \leq p_{i}^{infc} \leq P_{i2}$&Confine\\
    $P_{i2}\leq  p_{i}^{infc} \leq P_{i3}$&Quarantine\\
    $P_{i3}\leq  p_{i}^{infc} \leq 1$&Isolate\\
    \Xhline{1pt}
    \end{tabular}
\end{table}

\subsection{Avoiding extreme experiences} Similar to DURLECA where RL is used for epidemic control \cite{song2020reinforced}, it is possible to encounter extreme states or actions during the RL exploration in IDRLECA's training. This may severely impact exploration efficiency and result in local optimums. Inspired by DURLECA, we have a rule to avoid these extreme experiences:
\begin{itemize}
    \item The infection-increase threshold $I_{t}$: During the agent's exploration process, if the number of new infections on a certain day exceeds $I_{t}$, the current episode will be stopped and a large penalty will be given to the reward of the  agent.

\end{itemize}

\section{Experiments}
In this section, we conduct extensive experiments on four scenarios to answer the following research questions:

\begin{itemize}
    \item \textbf{RQ1:} Can IDRLECA minimize the number of infections and the cost of interventions?
    \item \textbf{RQ2:} Can IDRLECA be adapted to different scenarios?
    \item \textbf{RQ3:} How does IDRLECA compare to expert policies and PAPW winners?
\end{itemize}

\subsection{Experiment Setup}
In the following, we introduce more details about our experiment design.

\subsubsection{Simulation Enviroment}
We build a simulation environment mainly based on the PAPW Challenge~\footnote{PAPW 2020: https://prescriptive-analytics.github.io/. Simulator: https://hzw77-demo.readthedocs.io/en/round2/.}. The simulated disease has an $R0$ range from 2 to 2.5, which is similar to COVID-19~\footnote{World Health Organization. (2020, May 8). Report of the WHO-China Joint Mission on Coronavirus Disease 2019 (COVID-19). Retrieved May 8, 2020, from: https://www.who.int/docs/default-source/coronaviruse/who-china-joint-mission-on-covid-19-final-report.pdf}. The total simulation time is 60 days. Every individual has a pre-defined commute pattern.
To simulate a more practical EPC scenario, we add a new rule in the original simulator: all symptomatic patients should be sent to the hospital.

\subsubsection{Comparison Scenarios}
We define $t_{start}$ as the days to start epidemic intervention after discovering the first patient.

\begin{itemize}
        \item \textbf{Scenario-Default:} $N=11, M=10000, t_{start} = 1$. This scenario is to verify the EPC performance of IDRLECA in a ordinary epidemic scenario.
        \item \textbf{Scenario-Larger:} $N=98, M=10000, t_{start} = 1$. This scenario is to verify whether IDRLECA is suitable for scenarios with greater individual mobility.
        \item\textbf{Scenario-Changeable:} $N=11, M=10000, t_{start} = 1$. Compared with Scenario-Default, people's commute patterns are more changeable in this scenario. This scenario is to verify whether IDRLECA is applicable when there are greater differences in individuals' characteristics.
        \item\textbf{Scenario-Late:} $N=11, M=10000, t_{start} = 5$. Compared with Scenario-Default, this scenario starts intervention after 5 days of discoverying the first patient. This scenario is to verify the EPC performance of IDRLECA with a late intervention.
\end{itemize}

\begin{table*}[t]
    \centering
    \caption{Performance comparison in three scenarios when $t_{start}=1days$}
    \resizebox{0.99\textwidth}{!}
    {
    \begin{tabular}{c||ccc|ccc|ccc}
        \Xhline{1pt}
        & \multicolumn{3}{c|}{\textbf{Scenario-Default}} & \multicolumn{3}{c|}{\textbf{Scenario-Larger}}
        & \multicolumn{3}{c}{\textbf{Scenario-Changeable}}
        \\
        Method& I&Q &Score & I&Q &Score& I&Q &Score  \\
        \hline \hline 
        No Intervention&8289&123153.00&$>$10000&6588&92563.00&$>$10000&8115&113596.00&$>$10000\\
        Lockdown~\cite{hale2020variation}&\textbf{58}&294460.50&$>$10000&\textbf{56}&294508.50&$>$10000&\textbf{55}&294491.50&$>$10000\\
        Expert(0.01)&276&6997.50&3.75&204&9187.50&4.01&294&7837.00&3.99\\
        Expert(0.015)&319&8210.00&4.16&269&8404.50&4.03&328&8724.50&4.32\\
        \hline 
        Degree-Sample~\cite{watts1998collective}&1108&212940.00&$>$10000&1212&211146.50&$>$10000&943&212498.00&$>$10000\\
        Degree-Order~\cite{eubank2004modelling}&3557&120731.00&$>$10000&2302&92569.50&$>$10000&3133&119958.50&$>$10000\\
        \hline 
        GBM~\cite{c.elmohamed}&210&6408.21&3.42&177&4794.87&3.04&193&6091.76&3.31\\
        EITL~\cite{kimexpert}&220&7067.01&3.58&190&5640.15&3.22&205&7899.03&3.71\\
        \hline 
        HRLI~\cite{donghierarchical}&187&5689.79&3.22&183&4935.03
&3.08&187&7112.14&3.49\\
        \hline 
        IDRLECA&137&\textbf{3748.58}&\textbf{2.77}&170&\textbf{4606.17}&\textbf{2.99}&153&\textbf{4068.09}&\textbf{2.86}\\
        \Xhline{1pt}
    \end{tabular}
    }
\end{table*} 

\begin{table}[t]
    \centering
    \caption{Performance comparison in  Scenario-Late when $t_{start}=5days$}
    \resizebox{0.48\textwidth}{!}
    {
    \begin{tabular}{c||ccc}
        \Xhline{1pt}
        & \multicolumn{3}{c|}{\textbf{Scenario-Late}}
        \\
        Method& I&Q &Score  \\
        \hline \hline 
        No Intervention&8040&119175.00&$>$10000\\
        Lockdown~\cite{hale2020variation}&70&274364.50&$>$10000\\
        Expert(0.01)&340&8985.50&4.43\\
        Expert(0.015)&323&8388.00&4.22\\
        \hline 
        Degree-Sample~\cite{watts1998collective}&2091&195949.00&$>$10000\\
        Degree-Order~\cite{eubank2004modelling}&3331&115858.00&$>$10000\\
        \hline 
        GBM~\cite{c.elmohamed}&304&7808.13&4.02\\
        EITL~\cite{kimexpert}&291&8193.50&4.06\\
        \hline 
        HRLI~\cite{donghierarchical}&270&7197.86&3.77
        \\
        \hline 
        IDRLECA&193&5061.64&3.13\\
        \Xhline{1pt}
    \end{tabular}
    }
\end{table}

\subsubsection{Evaluation Metrics} 

\begin{itemize}
\setlength{\itemsep}{0pt}
\setlength{\parsep}{0pt}
\setlength{\parskip}{0pt}
        \item \textbf{$I$:} The total number of infected people in all simulation days. It is used to measure the effectiveness of EPC strategies in suppressing infections.
        \item \textbf{$Q$:} The aggregated mobility interventions defined in Section 2. To have a fair comparison with PAPW winners, we set $\lambda_{h}=1$, $\lambda_{i}=0.5$, $\lambda_{q}=0.3$ and $\lambda_{c}=0.2$ , which are the same with the setting in the PAPW Challenge.
        \item\textbf{$Score$:} The social cost of epidemic control policy which is defined in Section 2. We set $\theta _{I}=500$ and $\theta _{Q}=10000$, which are the same with the setting in the PAPW Challenge. 
\end{itemize}


\subsubsection{Comparison Baselines}  

We set up 4 expert baselines to simulate EPC strategies in the real world:

\begin{itemize}
        \item $No$ $Intervention$: No intervention at all.
        \item $Lockdown$~\cite{hale2020variation}: Lockdown the city for successive 60 days.
        \item $Expert(0.01)$ and $Expert(0.015)$: Baselines based on the infection probability model. Isolate individuals whose infection probability is higher than a given threshold.
\end{itemize}

We compare DIRLECA with two baselines commonly used in epidemic research:

\begin{itemize}
        \item $Degree-Sample$~\cite{watts1998collective}: If the number of an individual's acquaintances $n$ is more than 4, isolate the individual with a probability $(n-4)/n$.
        \item $Degree-Order$~\cite{eubank2004modelling}: Count the number of contacts of an individual in the past 5 days. Select the top $30\%$ for isolation.
     
\end{itemize}

We  compare $IDRLECA$ with PAPW winners:
\begin{itemize}
        \item $GBM$ \cite{c.elmohamed}:
         a  baseline for epidemic intervention by predicting individual health states, which strikes a balance between precision and recall.
        \item $EITL$ ~\cite{kimexpert}:
        a heuristic baseline that adjusts the epidemic strategy through a heuristic algorithm, which based on evaluating the intervention action effectiveness and  understanding resulting patterns and interpret causality.
        \item $HRLI$ ~\cite{donghierarchical}:
        a state-of-the-art RL baseline combining individual prevention with regional control.
\end{itemize}

\subsection{Results Analysis}  
We compare $IDRLECA$ with all the baselines when $t_{start}=1 day$. Table 2 shows our main results. $IDRLECA$ is  better than all baselines in three scenarios in metric Score. For instance, compared with the best baseline  $HRLI$ in Scenario-Default, our method can reduce the number of infected persons by $26.73\%$ and the cost of mobile intervention by $34.12\%$.

In the four expert baselines, we can find that $No$ $Intervention$  will aggravate the spread of infectious diseases and eventually lead to the paralysis of the medical system. The other three expert baselines can  limit the spread of epidemic to some extent. However, these strategies have paid huge mobile intervention costs in order to reduce the number of infections, thereby greatly increasing the total social cost. For example, $Lockdown$ is a common method in our real life when dealing with epidemic, which achieves best performance in  minimizing infections at the expense of the maximum mobile intervention cost.
Compared with the four expert baselines, $IDRLECA$ can minimize the infections and  retain large amounts of mobility at the same time. 

Compared to $GBM$, $EITL$ and two  baselines  commonly used in epidemic research, $IDRLECA$ performs better than them mainly because it considers more individual characteristics and the long-term impact of current actions when making decisions.

Compared to  $HRLI$, we find that $IDRLECA$ outperforms them
in all metrics which may be because the  GNN in our method models the contact between individuals and  estimates individual infection risks through contact. $IDRLECA$ can find hidden infections through GNN and thus be able to stop the spread of epidemic quickly at minimal mobility intervention cost, which  will be verified in Case Study.

From the results of  Scenario-Larger and Scenario-Changeable, we can find $IDRLECA$ can still guarantee the minimum number of infections and mobility intervention costs in more changeable and flexible scenarios.

Late intervention to an epidemic is very common in the real world. An effective control strategy should be able to stop the spread of the epidemic in time with the least cost of mobility intervention in the case of late intervention. 
We perform our experiment in Scenario-Late, and the results are shown in Table 3.  The results show that our method performs best in metric Score compared with other baselines in the case of late intervention.

In Figure 3, we compared the number of infections and the cost of mobility intervention between $IDRLECA$ and the best baseline method HRLI in Scenario-Late with $t_{start}=5days$ within 60 days. It can be found that our method can not only stop the spread of epidemic diseases faster, but also reduce the cost of intervention during the peak period of the epidemic.

In order to verify the effectiveness of our method for individual epidemic prevention and control, we randomly select 100 individuals in Scenario-Default, and draw a heat map of the  infection probability change within 60 days in Figure 4. It can be found that the infection probability of 100 people reaches its peak in about 15 days, but soon under the influence of intervention measures by $IDRLECA$, the probability of infection is soon reduced to 0 around the 40th day. 

\subsection{Case Study}
In order to verify the effectiveness of our method in individual prevention and control, we conduct two case studies.

\textbf{Evaluating individual intervention:} To verify the specific effects of our method on individual intervention, we draw the infection probability of a person  and the  changes in prevention and control measures within 60 days of Scenario-Default in Figure 5. It can be found that our method is very sensitive to the action control of different infection probabilities and can effectively reduce the risk of infection.

\textbf{Finding hidden infections:} In order to verify whether our method can discover hidden infections, we used IDRLECA to output actions to individuals with ID 927 and 959 on the 20th day of Scenario-Default: quarantine and confine. However, the infection probability of these two individuals is 0.004 and 0.34 respectively, whose numerical order is exactly opposite to the prevention and control level. We further found that the first person had more contacts and acquaintances in the past five days than the second person. This is because the infection probability calculated in Sec 3.1 only considers the impact of the current discovered infections and simplifies the spread of the epidemic by individual contacts. Our GNN models the contact between individuals and can estimate the individual's potential risk of infection and the ability to potentially infect others. Therefore, although the first person is relatively low in the probability of infection, our model takes into account the  infection risk which measures the harm and risk of secondary transmission of potentially infected individuals, so more strict measures are taken for the first person. Two days later, the first person was detected as infected during the intervention period, which also verifies our findings.

\begin{figure}[h]
    \centering
    \includegraphics[width=0.4\textwidth]{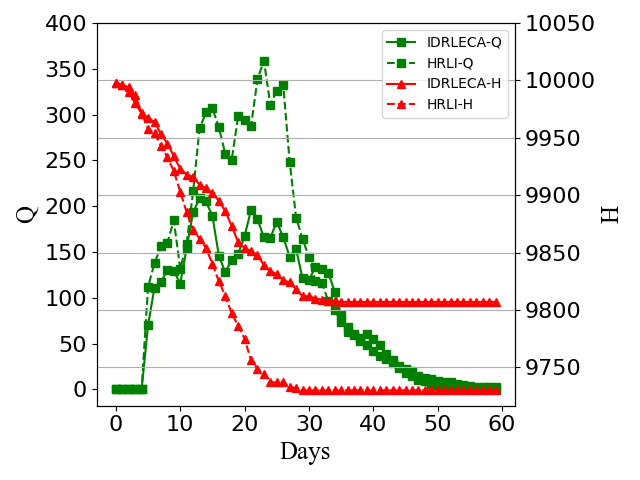}
    \caption{$Q$(the aggregated mobility interventions defined in Section 2) and $H$(the number of healthy people) are changing over time.} 
\end{figure}

\begin{figure}[h]
    \centering
    \includegraphics[width=0.4\textwidth]{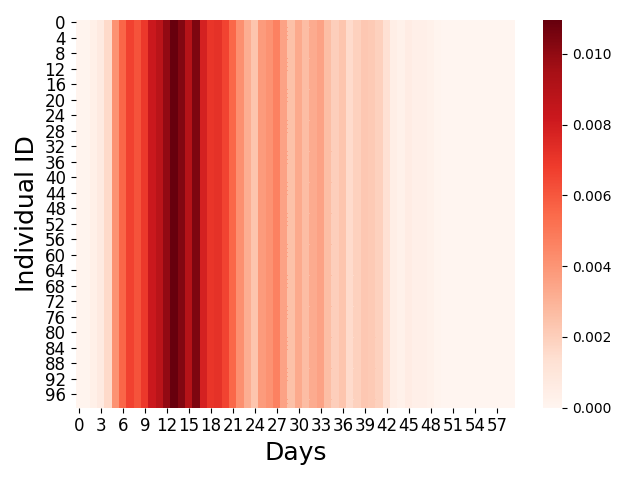}
    \caption{Change in infection probability of 100 individuals within 60 days(Scenario-Default).} \label{fig:penalty_50}
\end{figure}

\begin{figure}[h]
    \centering
    \includegraphics[width=0.4\textwidth]{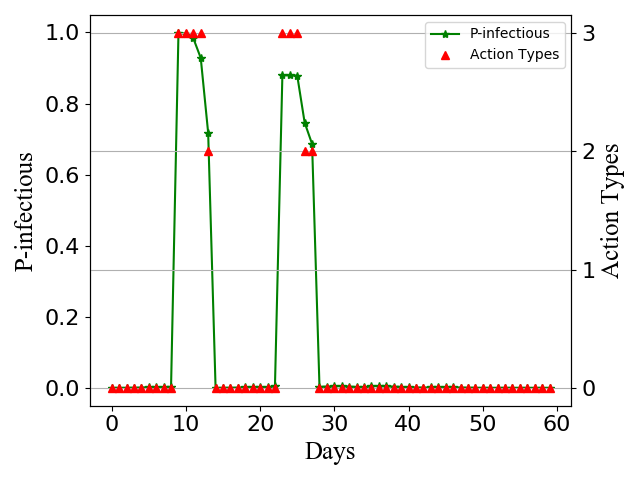}
    \caption{The relationship between infection probability and intervene action types.} 
    \label{fig:penalty_100}
\end{figure}

\subsection{Ablation Study}
To evaluate the effectiveness of our proposed  Individual Contact GNN and RL exploration strategy(Avoiding extreme experiences), we take ablation study in this section. We select three baselines and perform experiments in two scenes. No Intervene is the baseline of the blank control. RL-NoGraph and RL-NoEP denote removing GNN and RL exploration strategy(Avoiding extreme experiences) compared with IDRLECA. The results in Table 4 show that removing the GNN network structure will make it difficult for RL to find hidden infections, which will increase the number of infections and the cost of prevention and control. The removal of the exploration strategy(Avoiding extreme experiences) will make it hard for RL to further reduce the number of infections and the cost, falling into a local optimum. Compared with RL-NoGraph and RL-NoEP, IDRLECA  can better find hidden infections with the help of GNN, and ensure reasonable and effective exploration under the exploration strategy, so that it can learn better results.
\begin{table}[h]
    \centering
    \caption{Ablation study}
    \resizebox{0.48\textwidth}{!}
    {
    \begin{tabular}{c||ccc|ccc}
        \Xhline{1pt}
        & \multicolumn{3}{c|}{\textbf{Scenario-Default}} & \multicolumn{3}{c}{\textbf{Scenario-Default($t_{start}=5days$)}}\\
        Method& I&Q &Score & I&Q &Score  \\
        \hline \hline
        No Intervene&8289&123153.00&$>$10000&8040&119175.00&$>$10000\\
        \hline 
        RL-NoGraph&200&6606.32&3.42&313&7699.32&4.03\\
        RL-NoEP&192&5779.26&3.25&285&6678.26&3.82\\
        \hline 
        IDRLECA &137&3748.58&2.77&193&5061.64&3.13\\
        \Xhline{1pt}
    \end{tabular}
    }
    
\end{table}

 \section{Related Works}

\subsection{Individual-based Infection Simulation and Control Model:}
Individual-based Infectious Diseases Model(IBIDM)  is an epidemiology model that has emerged in recent years \cite{milne2008small}. Compared with traditional infectious disease models, IBDIM can reflect the heterogeneity of individuals and reflect individual-level behavior dynamics, thus more precisely reflect the spread of the epidemic. IBIDM models each individual as a unit, and measures the contact relationship between individuals through social contact network. The Los Alamos Laboratory in the United States has developed an individual-based infectious disease simulation tool, called EpiSimS system, which can effectively simulate the spread, prevention and control of the epidemic based on individual characteristics \cite{mniszewski2008episims}. Later, some researchers propose Epifast to simulate the spread of Ebola in West Africa, which has higher prediction accuracy and simulation preciseness than traditional methods \cite{bisset2009epifast}. There are also many researches related to epidemic control based on these epidemic simulation.
 \cite{augustine2020economy} studies the trade off between  spread of COVID-19 and economic impact and proposes some mechanisms based on group scheduling to  strike a balance between epidemic control and economic development. \cite{eubank2004modelling,park2018strength,watts1998collective}  regard individuals as nodes of the graph, and the connections between individuals as edges, and find the individuals who need to be isolated through graphs. \cite{wu2020individual} introduces mean-field models and complex networks to solve the individual prevention and control of the epidemic.

However, current researches are hard to to effectively extract the status of individuals and their strategies are often unable to cope with various scenarios and conditions. They often only pay attention to the current prevention and control effects, and do not care about the long-term impact of the current decision-making. Therefore, we develop IDRLECA which considers not only how to track and control the infectious and  asymptomatic
based on the infection status of individuals, but also how to achieve better epidemic prevention while minimizing economic losses in the long term. 

\subsection{Graph Neural Network for Individual Contacts:}
Graph Neural Networks (GNN) are mainly used for node prediction, link prediction and graph prediction tasks. Node prediction refers to predicting the type of a given node~\cite{kipf2016semi, liu2018heterogeneous}. 
Link prediction means predicting the connection status of two given nodes~\cite{zhang2018link, ying2018graph}. 
Graph prediction aggregates all node features in the graph as the graph feature, and then classifies the type of the graph based on it~\cite{bacciu2018contextual, geng2019spatiotemporal}. There are some commonly used GNN methods. GCN uses the adjacency matrix of nodes as input to learn the relationship between nodes~\cite{kipf2016semi}. GAT introduces an attention mechanism on the basis of GNN~\cite{velivckovic2017graph}. GraphSage learns node relationships by aggregating information from neighbor nodes~\cite{hamilton2017inductive}.

However, current GNN methods lack a framework to model the spread of epidemic  between individuals over a dynamic graph. Therefore, we propose a novel GNN structure to characterize the epidemic-spreading between individuals, whose nodes and edges  represent the state features of individuals  and contacts between individuals respectively.

\subsection{PPO:} PPO algorithm is a new type of policy gradient algorithm and has been applied in many aspects. \cite{schulman2017proximal} proposes that PPO strikes a balance between implementation simplicity, sample complexity and difficulty of tuning and  achieves good results in many games. \cite{berner2019dota,ye2020mastering} proves that PPO can perform well in solving some problems with  large-scale and complex state and action space. 

Since the PPO algorithm has good stability and adaptability, and can achieve good results in large-scale state and action space problems, we choose PPO as  RL algorithm in our problem.
\section{Conclusion}
In this paper, we propose IBRLECA that employs a novel GNN and RL approach to minimize infections as well as the mobility intervention cost  in EPC. The proposed GNN can estimate the spread of the virus through contacts between individuals. The training of IBRLECA is guided by a specially designed reward. We design and impose a constraint for control-action selection that eases its difficulty and further improve 
exploration efficiency.   
Extensive experiments are conducted on different scenarios to show the effectiveness of our proposed method.
\section{Appendix}
To help reproduce the results, here we present the details of the simulator\footnote{PAPW 2020: https://prescriptive-analytics.github.io/. Simulator: https://hzw77-demo.readthedocs.io/en/round2/.} and experiment settings.

\subsection{Introduction of Simulator}
The simulator contains a human mobility model and a disease transmission model. 
The simulator uses these two models to simulate individuals' movements and the spread of the epidemic among individuals. The two models are briefly introduced below:

\noindent\textbf{Human Mobility Model:}
The human mobility model simulates individual mobility in a city of $N$ areas with $M$ people. Each area is assumed to belong to one of the three categories: working, residential, and commercial. 
An individual is associated with two fixed areas: a residential area and a working area.  
We assume that an individual has different modes of mobility during weekdays and weekends. On weekdays, an individual will move from his/her residential area to his/her working area. After work, he/she may visit a nearby commercial area and then will return to his/her residential area. On weekends, an individual will visit a random commercial area.  After that, he/she will return to the residential areas.

\noindent\textbf{Disease Transmission Model:}
The disease can transmit from an infected individual through acquaintance contacts and stranger contacts. Contacts happen among people within the same region. The infection probabilities of contact with acquaintances and strangers are $P_{c}$ and  $P_{s}$, respectively. The disease transmission is simulated every hour.
\subsection{Experiment Setting}
We set the infection probabilities of contacting with strangers $p_{s}=0.01$  and infection probabilities of contacting with  acquaintances  $p_{c}=0.05$. The estimated $R0$ is 2-2.5. For the extreme-experience policy, we set $Q_{t}=250$. For $Score$, we set $\lambda_{h}=1$, $\lambda_{i}=0.5$, $\lambda_{q}=0.3$ and $\lambda_{c}=0.2$ , which are the same with the setting in the PAPW Challenge. For the reward and $Score$, we set $\theta _{I}=500$ and $\theta _{Q}=10000$.

In the training process, the beginning state of an episode is random every time. We train IDRLECA for 200,000 steps, using Adam optimizer with learning rate 0.0001. During testing, the initial setting is fixed in both IDRLECA and the baseline methods. Taking into account the randomness of the simulator, we compared the average results of all methods tested with three random seeds.

\subsection{Privacy and implementation issues}

\noindent\textbf{Privacy issues:}
Each area's visited history, the total number of people visiting a particular area and person-person relationship\cite{xu2016mining}can be obtained by individuals' trajectories. In practical system, user anonymity can be used to reduce the risk of privacy leakage for the individual trajectories and the health history data. 

\noindent\textbf{Implementation issues:}
Our system runs on a central server instead of individuals' smartphones and usually the policymaker has the ability to collect the data needed in our model. Besides, some recent techniques like Apple and Google APIs\footnote{https://covid19.apple.com/contacttracing} can be used to collect data without using private information. 
In the practical implementation of our method, distributed servers and federated learning can be used to protect privacy. The city will be divided into small areas. In each area we have a distributed server that receives encrypted data from smartphones and conducts federated learning with the central server. After training, each distributed server pulls the model from the central server and send reminders to users' smartphones. 
\label{sec:con}


%

\ifCLASSOPTIONcompsoc
  \section*{Acknowledgments}
\else
  \section*{Acknowledgment}
\fi

This work was supported in part by The National Key Research and Development Program of China under grant 2018YFB1800804, the National Nature Science Foundation of China under U1936217,  61971267, 61972223, 61941117, 61861136003, Beijing Natural Science Foundation under L182038, Beijing National Research Center for Information Science and Technology under 20031887521, and research fund of Tsinghua University - Tencent Joint Laboratory for Internet Innovation Technology.

\ifCLASSOPTIONcaptionsoff
  \newpage
\fi



\bibliographystyle{IEEEtran}
%



\bibliography{bibliography.bib}

%






\end{document}